\def\year{2022}\relax
\documentclass[letterpaper]{article} 
\usepackage{aaai22}  
\usepackage{times}  
\usepackage{helvet}  
\usepackage{courier}  
\usepackage{graphicx} 
\usepackage{natbib}  
\usepackage{caption} 
\DeclareCaptionStyle{ruled}{labelfont=normalfont,labelsep=colon,strut=off} 
\usepackage{amssymb}
\usepackage{multirow}
\usepackage{algorithm}
\usepackage{algorithmic}
\usepackage{subfigure}
\usepackage{amsmath}
\usepackage{xcolor}
\newcommand{\ie}{\emph{i.e.}}
\newcommand{\eg}{\emph{e.g.}}
\newcommand{\etal}{{\em et al.}}
\newcommand{\etc}{{\em etc.}}
\newcommand{\system}{Vulcan}
\usepackage{newfloat}
\usepackage{listings}
\floatstyle{ruled}
\newfloat{listing}{tb}{lst}{}
\floatname{listing}{Listing}
\title{\system{}: Solving the Steiner Tree Problem with Graph Neural Networks and Deep Reinforcement Learning}
\author{
    Haizhou Du\textsuperscript{\rm 1},
    Zong Yan\textsuperscript{\rm 1},
    Qiao Xiang\textsuperscript{\rm 2},
    Qinqing Zhan\textsuperscript{\rm 1}
}
\affiliations{
    \textsuperscript{\rm 1}Shanghai University of Electric Power \\
     \textsuperscript{\rm 2}Xiamen University 

}

\begin{document}

\maketitle

\begin{abstract}
Steiner Tree Problem (STP) in graphs aims to find a tree of minimum weight in the graph that connects a given set of vertices. It is a classic NP-hard combinatorial optimization problem and has many real-world applications (\eg, VLSI chip design, transportation network planning and wireless sensor networks).
Many exact and approximate algorithms have been developed for STP, but they suffer from high computational complexity and weak worst-case solution guarantees, respectively. Heuristic algorithms are also developed. However, each of them requires application domain knowledge to design and is only suitable for specific scenarios.
Motivated by the recently reported observation that instances of the same NP-hard combinatorial problem may maintain the same or similar combinatorial structure but mainly differ in their data, we investigate the feasibility and benefits of applying machine learning techniques to solving STP. To this end, we design a novel model \system{} based on novel graph neural networks and deep reinforcement learning. The core of \system{} is 
 a novel, compact graph embedding that transforms high-dimensional graph structure data (\ie, path-changed 
 information) into a low-dimensional vector representation. Given an STP instance, \system{} uses this embedding to encode its path-related information and sends the encoded graph to a deep reinforcement learning component based on a double deep Q network (DDQN) to find solutions. In addition to STP, \system{} can also find solutions to a wide range of NP-hard problems (\eg, SAT, MVC and X3C) by reducing them to STP.
 We implement a prototype of \system{} and demonstrate its efficacy and efficiency with extensive experiments using real-world and synthetic datasets.

\end{abstract}

\section{Introduction}
\label{sec:introduction}

The Steiner Tree Problem (STP) in graphs is defined as the problem to find a minimum-weight tree in a graph to connect a given set of vertices. As a representative graph combinatorial optimization problem, it was first proposed in \cite{dreyfus1971steiner} and proved to be NP-hard. STP and its variants have many applications in real world, such as very large scale integration (VLSI) chip design \cite{grotschel1997steiner}, multimodal transportation network planning \cite{van2002design}, sensor network routing protocol design \cite{bern1988network}, and analysis of social network \cite{kasneci2009star}.

Given the many applications of STP in real world and its NP-hardness, many algorithms have been developed for solving STP. For example, \cite{dreyfus1971steiner} designed exact algorithms to find optimal solutions to STP with a complexity of O($3^{\left|S\right|}poly(n)$), where $n$ is the number of vertices of the graph and $S$ is the set of terminals. \cite{KouA} designs approximate algorithms that finds solutions to STP with an approximation ratio of $2-2/t$, where $t$ is the number of leaves in the optimal Steiner Tree. In addition, people also leverage domain knowledges to develop different heuristic algorithms to solve STP in different applications \cite{Luyet2007An,kasneci2009star,van2002design}. Although substantial progress has been made in solving STP, different algorithms have different limitations. Specifically, exact algorithms suffer from the non-polynomial computation complexity; approximate algorithms suffer from the poor worst-case guarantee in large-scale STPs; and heuristic algorithms heavily rely on domain expertise of the targeted applications to be efficient and useful, \ie, algorithms for one STP application will not work well for a different application.

In this paper, we investigate the feasibility and benefits of applying machine learning theories and techniques to solving STP. Our main motivation comes from (1) it is  recently observed that instances of the  same  NP-hard  combinatorial  problem  may  maintain the same or similar combinatorial structure, but mainly differ  in  their  data \cite{khalil2017learning}, and this observation has been long overlooked when designing algorithms for NP-hard problems; and (2) 
{existing studies have shown success in solving NP hard problems by representing the problem as a graph and applying machine learning techniques}, such as the satisfiability (SAT) problem, the maximal vertex cover (MVC) problem and the maximum independent set (MIS) problem \cite{khalil2017learning,li2018combinatorial,peng2020graph}. Different  in  details,  they share a common idea: reduce the problem-in-interest (\eg, the SAT problem) to a graph problem (\ie, the MIS problem), use graph embedding to encode the problem information into a d-dimensional vectors, and feed the embedded graph to corresponding learning modules (\eg, graph convolutional networks and reinforcement learning) to find solutions. 

Despite the progress of these studies, however, applying machine learning to solve the STP is still non-trivial. The fundamental challenge is that the graph embedding of these studies is too primitive to compactly encode node-related and path-related information, which is high-dimensional data in graph and crucial for finding solutions to STP. Specifically, in STP, there are two types of vertices: the given set of vertices to be connected (called the \textit{terminals}), and the other vertices in the graph (called \textit{non-terminals}). In contrast, problems focused by existing studies (\eg, MVC and MIS) only have one type of vertices in graph.  In addition, to connect terminals in STP, different paths need to be searched, concatenated and compared. In contrast, solving problems like MVC and MIS do not need need path information in graph.

To this end, we design a novel model \system{} based on a novel Graph Neural Networks (GNN) and deep reinforcement learning. The core of \system{} is 
 a novel, compact graph embedding, which transforms high-dimensional graph structure data (\ie, node related and path-related information) into a low-dimensional vector representation. As such, given an STP instance, \system{} can encode its node/path-related information compactly and sends the encoded graph to a deep reinforcement learning component based on double deep Q network (DDQN) \cite{van2015deep} to find solutions. In addition to STP, \system{} can also find solutions to a wide range of NP-hard problems (\eg, SAT, MVC and X3C) because they can be reduced to STP \cite{promel2012steiner,hartmanis1982computers}.

The \textbf{main contributions} of this paper are as follows:
\begin{itemize}
\item To the best of our knowledge, \system{} is the first approach that uses deep reinforcement learning to solve the STP; 

\item As the core of \system{}, we propose a novel graph embedding moddel, which can transform high-dimensional graph structure data into low-dimensional vector representation by capturing path extraction information for STP;  

\item We reformulate the STP as a sequential decision-making process, and design a DDQN-based deep reinforcement learning component in \system{} to solve STP, as well as a wide range of other NP-hard combinatorial problems by reducing them to STP;

\item  We implement a prototype of \system{} and demonstrate its efficiency and efficacy with extensive experiments using both synthetic datasets and real-world datasets.
\end{itemize}

The rest of the paper is organized as follows. The related work is discussed in Section \ref{sec:related}. In Section \ref{sec:preliminares}, we formally describe some preliminaries for STP. The specific design ideas of \system{} will be proposed in Section \ref{sec:design}. 
Experimental methodology and results are given in Section \ref{sec:evaluation}.  Finally, we give concluding remarks in Section  \ref{sec:conclusion}.

\section{Related Work}
\label{sec:related}
The STP in graphs is a well known NP-hard problem and many algorithms have been proposed since 1972 \cite{karp1972reducibility}. We classify these approaches into two categories: traditional methods and simple neural network methods. In addition, we introduced the related work of GNN and reinforcement learning on combinatorial optimization problems. Although they are other np-hard problems, we are inspired by them.

\subsection{Traditional methods for STP in graphs}
According to the characteristics of the problem, these traditional methods directly give solutions that are often effective. Kou L \etal~\cite{KouA} proposed a fast algorithm for the STP by constructing a minimum spanning tree and 
{looking for other shorter paths to replace some edges in the minimum spanning tree}. Luyet L \etal~\cite{Luyet2007An} designed an ant algorithm called ANT-STP for STP in graphs which is inspired by the information exchange between natural ants. On a weighted undirected graph, the sum of the weighted edges taken by each ant is taken as the path, so each ant has its unique path information. 
They interact with each other at nodes and finally form a complete trail system. 
{Esbensen \etal~\cite{esbensen1995computing} presented a new genetic algorithm (GA) for the STP. The algorithm is based on a bitstring encoding of selected Steiner vertices and the corresponding STP is computed using a deterministic STP heuristic.} In addition, there is also a class of classical algorithms for solving STP in graphs called the branch-and-cut method \cite{ChopraSolving,KochSolving} that using robust Mixed integer programming (MIP) solvers with various Steiner-tree-specific features. But they are limited to the number of terminal nodes and perform well only for examples with a small number of terminals. To solve this problem, Iwata Y \etal~\cite{iwata2019separator} presented a novel separator-based pruning technique for speeding up a theoretically fast dynamic planning (DP) algorithm that can achieve good results on hundreds of terminal nodes. Although the method is very competitive with the branch-and-bound algorithm, they have one common feature: the algorithm design is extremely complex.

\subsection{Simple neural network methods for STP in graphs}
Compared to the complex design of traditional algorithms, a simple neural network model consists of only several hidden layers. Zhongqi J \etal~\cite{Jin1993Using} proposed an algorithm based on neural computing, which is very simple and easy to realize. Pornavalai C \etal~\cite{Pornavalai1996Neural} proposed a modified Hopfield Neural Network to solving STP in graphs. The proposed model can integrate path information to dynamically adjust path selection to achieve minimum cost. 
Compared to previous neural networks, their results have improved. However, even for small-scale graphs, the optimal solution could not be found every time. More importantly, the neural network cannot understand the  graph's topology, so it cannot be extended to other graphs.

\subsection{GNN for combinatorial optimization}
In recent years, the success of deep learning methods has led to increasing attention being given to combinatorial optimization problems. 

For example, Khalil E \etal \cite{khalil2017learning} 
presented an end-to-end deep learning model for challenging combinatorial optimization problems on graphs, such as Minimum Vertex Cover (MVC), Maximum Cut (MC) and Traveling Salesman problems (TSP). The model is composed of deep graph embedding and unique reinforcement learning. They 
used structure2vec (S2V) \cite{ribeiro2017struc2vec} to embed the graph that trained a graph embedding network to compute a low-dimensional feature embedding for each node. Then they used Q to represent the parameterized nodes' value and finally applied it to reinforcement learning. 
{Compared with the heuristic algorithm, their method reduces the complex artificial design.
Graph convolutional network (GCN) as a kind of representational technique can capture the neighboring nodes' information.} Li Z \etal \cite{li2018combinatorial} used GCN \cite{defferrard2016convolutional,kipf2016semi} as the graph embedding method for the Maximal Independent Set (MIS) problem. The GCN was trained to predict the probability that each node belongs to the optimal solution, and then greedily constructs the optimal solution. Finally, a parallelized tree search procedure was used to generates a large number of candidate solutions, one of which is chosen after subsequent refinement.
Velikovi P \etal \cite{velikovi2020neural} focused on learning in the space of algorithms that trained several state-of-the-art GNN architectures to imitate individual steps of classical graph algorithms, although they are not NP-hard problems. This is a fascinating, and experiments have verified its effectiveness on graphs by learning several classic algorithms. 

Compared to manual algorithm designs, GNN-based methods can automatically identify distinct features
from training data and adapt to a family combinatorial optimization problems.
\subsection{Reinforcement learning for combinatorial optimization}
Recently, reinforcement learning methods, jointly with graph representation solve the  combinatorial problems by training an agent that takes the decision on the construction of the optimal path. In particular, the Maximum Cut~\cite{barrett2019exploratory}, Minimum Vertex Cover~\cite{khalil2017learning,song2020co}, Traveling Salesman~\cite{khalil2017learning,cappart2020combining}, Set Covering Problem~\cite{khalil2017learning}, Maximum Independent Set~\cite{cappart2019improving}, Maximum Common Subgraph~\cite{bai2020fast} and 4-moments portfolio optimization~\cite{cappart2020combining} are some fundamental problems have been solved by reinforcement learning. However, due to the complexity of STP, there still is not any solution using reinforcement learning to solve the STP.

\section{Preliminaries}
\label{sec:preliminares}
In this section, we first describe the STP in graphs and give the definition. Then we introduce prior knowledge about GNN and GNN-based graph embedding. Last, we describe a greedy algorithm of reinforcement learning on graphs. Symbols used in this paper see Table \ref{notation}.
\begin{table}[htbp]
\footnotesize
\centering
\caption{Symbols used in this paper}
\begin{tabular}{c|c}
\hline
\textbf{Symbols} & \textbf{Description} \\ \hline
$G$ & an undirected graph \\ \hline
$V$ & set of all vertices (nodes) \\ \hline
$E$ & set of all edges \\ \hline
$\omega$ & the edge weight of a pair of vertices \\ \hline
$T$ & set of terminals \\ \hline
$h_i^{t}$ & the $t$-th layer representation of vertex $i$ \\ \hline
$\widehat h_i^{t+1}$ & adjacency aggregation representation of vertex $i$\\ \hline
$N_i$ & the set of neighbors of vertex $i$ \\ \hline
$\theta$ & set of learnable parameters \\ \hline
$S$ & the partial solution \\ \hline
${S}^{*}$ & the vertices that can be added \\ \hline
${s}_{v},{t}_{v}$ & the state of vertex $v$ \\ \hline
${v}^{*}$ & the highest value vertex \\ \hline
$Q$ & the evaluation function \\ \hline
$C$ & the objective function \\ \hline
$K$ & K-nearest terminals \\ \hline
$X$ & matrix of shortest distance between vertices and terminals \\ \hline
$\mu_v$ & hidden vector representation of encoder network \\ \hline
${\mu_v}'$ & hidden vector representation of processor network \\ \hline
$r$ & the reward function \\ \hline
$v'$ & the vertex of the maximum Q value of the next state \\ \hline
$\gamma$ & discount rate \\ \hline

\end{tabular}
\label{notation}
\end{table}

\subsection{The STP in graphs}
\begin{figure}[htbp]
\centering
\includegraphics[scale=0.5]{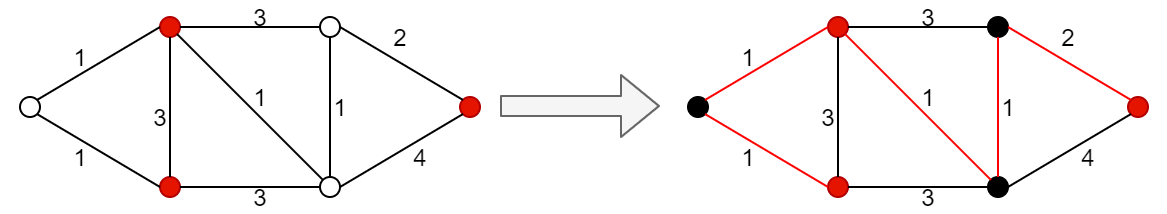}
\caption{Connecting all terminals(red vertices) with the shortest path(red edges) on the graph. The black vertices indicate the steiner vertices that belong to the shortest path. }
\label{STP}
\end{figure}

Given a graph with weighted edges, the STP aims to find a tree of minimum weight that connects a set of vertices. Formally, it is defined as follows.\\
\textbf{Definition.} \cite{karp1972reducibility} \textit{Given an undirected graph $G(V, E, \omega)$ with a set of $V$, a set of $E$ and $\omega$ : $ E \mapsto {R}^{+} $ the edge weight. For a set $ T \in V $, our problem consists of finding a subgraph of $ G$ with minimal cost that contains at least all vertices of a set $T$ (called terminals).} see Fig. \ref{STP}.

The STP is NP-hard that can be proved by reducing a wide range of NP-hard problems such as the satisfiability (SAT) problem, the maximal vertex cover (MVC) problem, the maximum independent set (MIS) problem, and the exact cover by 3-sets (X3C) to STP \cite{promel2012steiner,hartmanis1982computers,santuari2003steiner}.



\subsection{GNN}
\begin{figure}[htbp]
\centering
\includegraphics[scale=0.8]{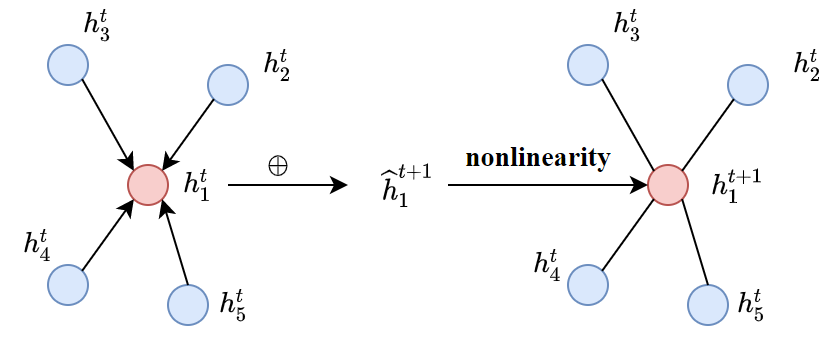}
\caption{Illustration of GNN iterative update process}
\label{GNN}
\end{figure}

GNN benefits a wide variety of graph analytics applications, often used in link prediction \cite{zhou2017scalable}, recommender systems \cite{ying2018graph}, knowledge graph \cite{wang2014knowledge,lin2017learning}, social networks alignment \cite{man2016predict,liu2016aligning}, etc. Although there are many variants of GNN, we describe a simple form that can cover the most central ideas. GNN iteratively updates
node representations from one layer to the other according to the formula:
\begin{equation}
\widehat h_i^{t+1}=\underset{j\in N_i}\oplus h_j^t
\end{equation}
\begin{equation}
h_i^{t+1}=\sigma(\theta^t \widehat h_i^{t+1})
\end{equation}
where $\oplus$ is an elementwise aggregation
operator, such as maximization, summation and averaging. Different aggregation methods have different effects. There can also be more complex methods, such as Graph Attention Networks (GAT) \cite{velickovic2018graph}, an attention mechanism enabling vertices to weigh neighbor representations during their aggregation. $h_i^{t+1}$ is a $P$-dimensional embedding representation of node $i$ at layer $t+1$, $j\in N_i$ is the set of nodes connected to node $i$ on the graph. $\theta^t$ is a learnable parameter, and $\sigma$ is a nonlinearity. For an intuitive illustration, see Fig. \ref{GNN}.

However, due to the two major limitations of GNN, it cannot express our problem. We take the model of GCN and tree search as an example \cite{li2018combinatorial}. On the one hand, the number of GCN layers is too shallow, and most of them do not exceed three layers. STP's terminals can be distributed at any location on the graph, and it is impossible for GCN to capture remote terminals information. On the other hand, GNN cannot handle heterogeneous graphs which have 
{two or more} types of vertices or edges. Terminals and non-terminals can be seen as two different types of vertices. In addition, tree search relies on high-quality labels, which is expensive for NP-hard problems, and tree search is a search process, so it is time-consuming in large graphs. 
{In contrast, reinforcement learning will more intelligent than search methods.}

\subsection{Graph embedding for combinatorial optimization} 
A graph has to be represented in numerical vectors to solve graph-based combinatorial optimization problems using machine learning. This process is called graph embedding. Graph embedding encodes graphs and vertices into low-dimensional vectors while preserving the structural properties of graphs. Among different graph embedding methods, GNN-based methods, which use neural network models specially defined on graphs, are the most widely used embedding for combinatorial optimization. We refer readers to \cite{peng2020graph} for a comprehensive survey of graph embedding. However, these embedding methods cannot compactly encode heterogeneous vertices and path information. Such information is high-dimension in nature and crucial for solving STP. For example, vertices in STP are categorized as terminal and non-terminal, and paths need to be searched, concatenated and compared to connect terminal vertices. As such, a fundamental challenge of using machine learning to solve STP is \textit{how to encode such high-dimension information compactly}.



\subsection{A Greedy Algorithm of reinforcement learning on Graphs} 
\label{greedy}
We will focus on a popular pattern for designing approximation and heuristic algorithms, namely a greedy algorithm that constructs a partial solution $S$ by gradually adding vertices and selecting vertices based on an evaluation function $Q$. 
Next, we express the greedy algorithm with a formulation.

A partial solution is represented as a list $ S=({v}_{1},{v}_{2}, \cdots,{v}_{|s|}), {v}_{i} \in V, {v}_{1} \in T $. Starting from a terminal, a vertex that outside the partial solution and connected to the partial solution is added each time. We use a set ${S}^{*} $ to represent vertices that can be added in the current state. In addition, we set a binary scalar $ s$ with each ${s}_{v}$ corresponding to a vertex $ v \in V $. If $ v \in S $, $ {s}_{v}=1 $, 0 otherwise. Similarly, we set a binary scalar $t$ with each $ {t}_{v} $ corresponding to a vertex $ v \in V $. If $ v \in T $, $ {t}_{v}=1 $, 0 otherwise.

A general greedy algorithm selects the maximum evaluation function $ Q(S,v) $, which depends on the current partial solution and the next vertex $ v$ to be added. Then, the partial solution $ S $ will be extended as:
\begin{equation}
S:=(S,{v}^{*}),{v}^{*}:={\arg\max}_{v \in {S}^{*}} Q(S,v)
\end{equation}
$ {v}^{*} $ represents the vertex with the highest value selected by the evaluation function $ Q(S,v)$ and then it is added to the partial solution $ S$  to generate a new state until all terminals are added.

The quality of the partial solution $ S$ can be defined by an objective function $ C$:
\begin{equation}
C(S,G) =  - \sum\limits_{i = 1}^{\left| S \right|} {\min \omega (u,v_i^*),u \in {S_i}} 
\end{equation}
The cost of adding a vertex is equivalent to the current partial solution $ {S}_{i} $ to the added node $ {v}_{i}^{*} $ by selecting the edge with the smallest weight. The termination criterion is activated when $ T \in S $. 
{Our goal is to maximize $ C(S,G) $ in the final state.}

\section{\system{} Approach}
\label{sec:design}
We present the design of \system{} in this section. Specifically, we first give an overview of the \system{} architecture in Section \ref{sec:overview}. Section \ref{sec:embedding}, gives the details on how our novel graph embedding mechanism extracts path information of  an STP instance and compactly encodes such high-dimension information into a low-dimension vector. We then describe in Section~\ref{sec:rl} how we construct a DDQN to find solutions for STP with such compact embedding.

\subsection{System Overview}
\label{sec:overview}

\begin{figure*}[htpb]
\centering
\includegraphics[width=\linewidth]{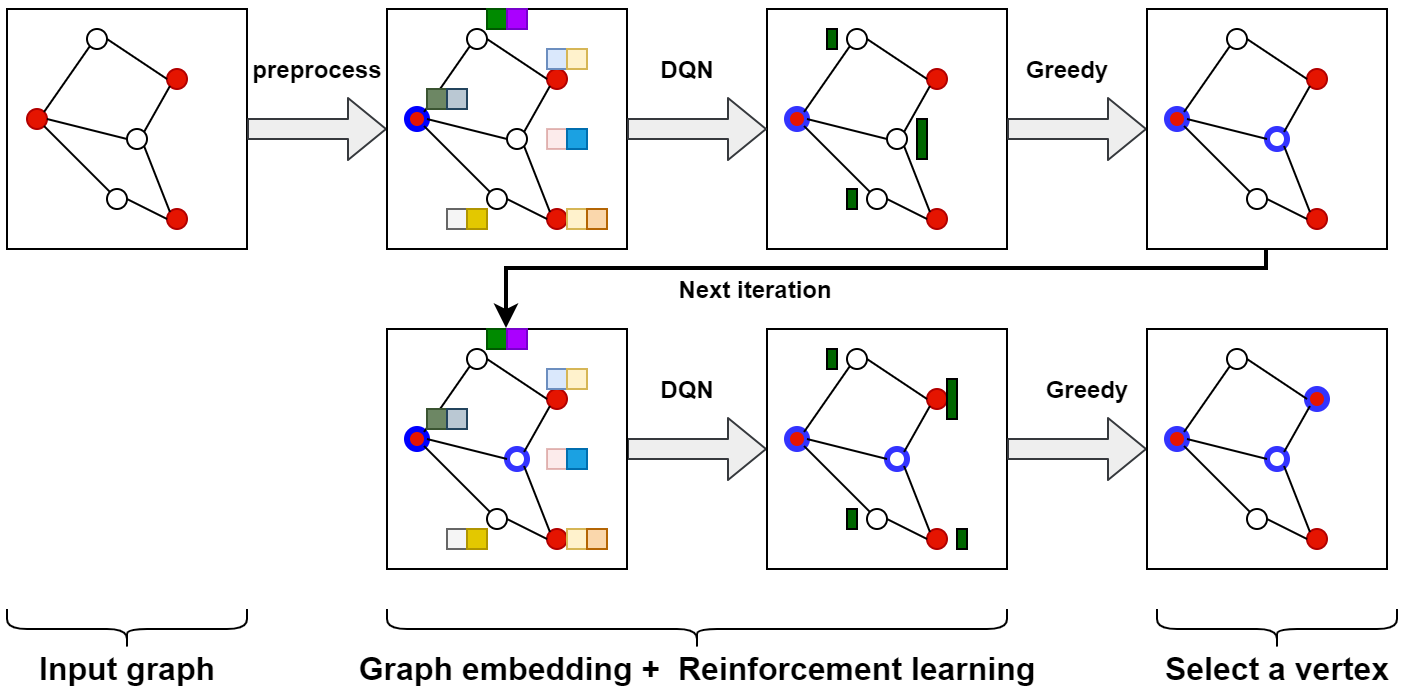}
\caption{The overview of \system{}}
\label{fig:flowchart}
\end{figure*}

Fig. \ref{fig:flowchart} presents the overview of the \system{} model. First, given an input graph (where red vertices represent terminals), \system{} obtains vertices information through a preprocessing process. It then uses a novel graph embedding to compactly encode  the vertices information and path-related information as vertex states (indicated as the colored bar next to the vertices). Next, \system{} feeds the embedded graph to the DDQN to train the score of optional vertices. (the higher the green bar, the higher the score) In the end, \system{} constructs the solution by greedily adding vertices with the highest scores. 

\subsection{Graph Embedding Model}
\label{sec:embedding}
In this section, we introduce the graph embedding methods that have been applied to solve graph-based combinatorial optimization problems. Most graph embedding methods have two stages. The first stage is preprocessing the input graph, and the second stage is training a machine learning model.
In solving graph-based combinatorial optimization problems using ML-based methods, a graph has to be represented in numerical vectors, which is known as graph embedding \cite{hamilton2017representation}.
\subsubsection{Preprocessing of Input Graph}
\label{Initial}
\system{} focuses on the vertices. The weight of each vertex depends on the
distribution of the terminals, and also represents the probability of belonging to
the Steiner vertices. Since STP is an NP-hard problem, the exact weight of each
vertex cannot be found, so we obtain the weight-related features of each vertex
on the graph by simple preprocessing. Through the analysis of greedy algorithms
on STP, when adding an adjacent vertex $v_i$ every time, the first thing we
should notice is the distance of the vertex $v_i$ to other terminals. We use a
matrix $X\in{\mathbb{R}^{|V| \times |T|}} $ to denote the shortest distance weight, with the $i$-th row and $j$-th column $x_{ij}$
as the shortest distance between the vertex $v_i$ and terminal $v_j$, where $i
\in |V|$ represents the number of vertices $V$ and $j \in |T|$ represents number
of terminals $T$. 

There are some problems with such a matrix. On the one hand, different graphs will have unequal numbers of terminals, which will lead to different dimensions of the matrix columns. On the other hand, for a certain vertex, not all terminal information is useful, which will also lead to a waste of storage space. The smaller the shortest distance from the terminal to the vertex $v_i$, the greater the influence on the vertex. On the contrary, the terminals far away from the vertex $v_i$ are not considered.

So we set a constant parameter $K$ to represent the number of the vertex to $K$-nearest terminals and update the matrix to $X \in {\mathbb{R}^{|V| \times |K|}} $. As for the appropriate value of $K$, we will explain it in the following experiment. Another point to consider is that when a terminal is added to the partial solution, the terminal will lose value. So $K$ should refer to those terminals that have not been added to the partial solution. Finally, when the number of terminals is less than $K$, we take the 0-filling operation. 
{Because as terminal vertices are added to the partial solution, the number of effective terminal vertices will gradually decrease.}

\subsubsection{Learning from the {Prepossessed} Results}
\begin{figure*}[htpb]
\centering
\includegraphics[width=\linewidth]{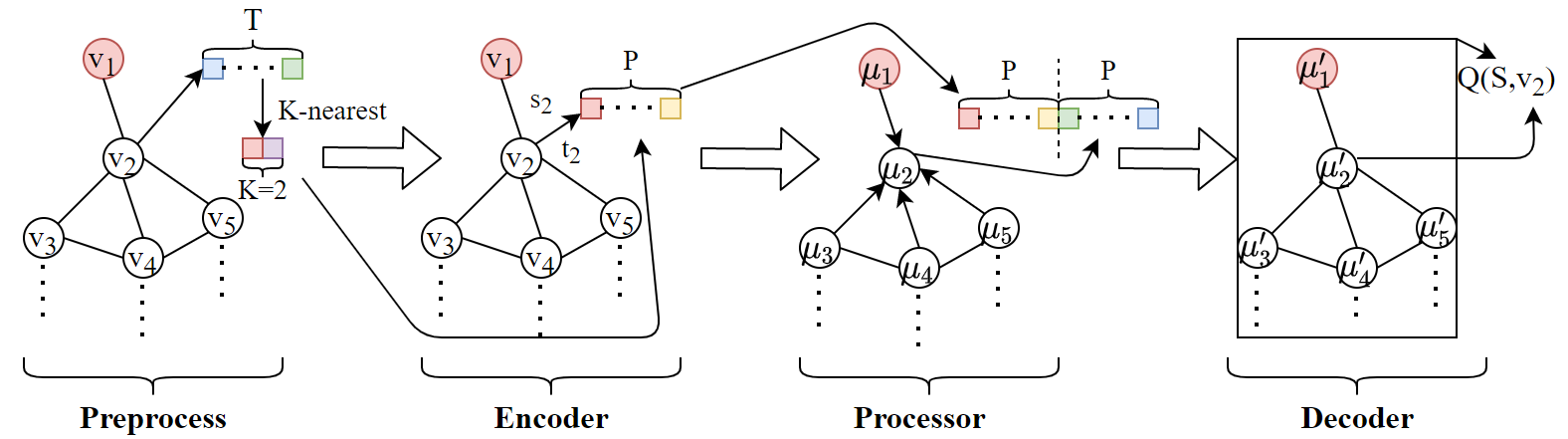}
\caption{The model of graph embedding. The color bars represent features that have different dimensions and the red vertex is terminal. The first step is to preprocess the input graph to obtain the initial vertex weights. Then the encoder network integrates the current vertex $v_2$ state and weight information to generate a P-dimensional hidden vector representation. Next, the processor network captures the changes between the vectors and splices them behind the P-dimensional vector. Finally, the decoder network considers the state of the entire graph and the current vertex $v_2$ to generate the value $Q$ of the vertex.}
\label{fig:embedding}
\end{figure*}

Since the STP is unstructured data, vertex's value in the context depends on many factors, such as graph structure, terminal position, vertex degree, etc. Our goal is that the evaluation function $Q$ can understand the current state and figure out the value when a new node is added in the context of such a graph. Towards that, we design an encode-process-decode architecture (see Fig. \ref{fig:embedding}) and then explain the purpose of each sub-network. 

First, we define an encoder network to integrate the partial solution $S$ and the initial vertex weight:
\begin{equation}
{\mu _v} = \mathsf{relu}({\theta _1}[{s_v},{t_v}] + {\theta _2}{x_v}),
\end{equation}
where $ {\theta _1} \in {\mathbb{R}^{p \times 2}},{\theta _2} \in {\mathbb{R}^{p \times K}} $ are the model parameters. $\mathsf{relu}$ is the rectified linear unit and $[\ ,\ ]$ denotes the connection operation. ${x_v}$ represents the initial vertex weight from preprocessing of input graph, which is passed to the encoder network after normalization. $ {\mu _v} $ is a P-dimensional vertex embedding. This part is to get global path information and the status of vertices.

Next, the vertex embedding  $\mu _v$ is processed by the processor network. The processor network updates a vertex hidden embedding ${\mu _v}'$ following a message-passing strategy from neighbors:
\begin{equation}
{\mu_v}' =l_\theta \mathsf{relu}\lbrack\mu_v,{\textstyle\sum_{u\in N(v)}}(\mu_v-\mu_u)\rbrack,
\end{equation}
where $N(v) $ represents the neighbors of vertex v and summation is a way to aggregate neighbor information. The subtraction operation is to capture the path information between a pair of vertex embeddings, and then we combine it with current vertex embedding ${\mu _v}$. In order to make the nonlinear transfer more powerful, we added several fully connected layers $l_\theta$ after the combination operation. This part is to collect local path change information while retaining the global path information. 

Finally, decoder network uses the state of the entire graph $ \sum\nolimits_{u \in V} {{\mu _u}'} $ and the hidden embedding $ {\mu _u}' $ to define the evaluation function $Q$, then to parameterize $ Q(S,v,\theta) $:
\begin{equation}
Q(S,v;\theta ) = \theta_3^\tau \mathsf{relu}([{\theta _4}\sum\nolimits_{u \in V} {{\mu _u}',{\theta _5}{\mu _v}'])},
\end{equation}
where ${\theta _3} \in {\mathbb{R}^{2p}},{\theta _4},{\theta _5} \in {\mathbb{R}^{p \times p}}$. $u \in V$ refers to all vertices in the graph. 
The graph embedding we designed is suitable for different types and sizes of graphs and all parameter sets are learnable. So we can train on small-scale graphs, which guarantee the speed of convergence. Then $\theta$ retain the training parameters and expand them to test on large graphs, and an excellent approximation ratio can still be obtained. It will be discussed in a later experiment.


\subsection{Double Deep Q Network Construction}
\label{sec:rl}
Since the STP is an NP-hard problem, we do not use supervised learning for optimization \cite{dai2016discriminative}, because obtaining the label of each vertex not only requires other heuristic algorithms but also accuracy cannot be guaranteed. 
{STP is a decision-making problem, and reinforcement learning is very suitable for solving such problems. At the same time,} the evaluation function $ Q(S,v,\theta) $ constructed by Section \ref{sec:embedding} can be naturally extended to reinforcement learning \cite{sutton2018reinforcement}. We use value-based reinforcement learning represented by DDQN, which has higher sampling efficiency compared to policy-based methods. To learn different types and sizes of graphs for the function $ Q(S,v,\theta) $, we define the following six aspects: state space, transition, action space, rewards, policy, and termination.
\begin{itemize}
\item \textbf{State Space:} 
{The state can be represented by the embedded vertex vector through graph embedding technology, including partial solution $S$ and other vertices on a graph $ G$.} We have used graph embedding to express such a state as an evaluation function $ Q(S,v,\theta) $, 
{such a model can identify the terminals and capture every state change on the graph when a vertex is added.}
\item \textbf{Transition:} The state transition of STP on the graph is deterministic, and there is no probability problem. A vertex $ v$ is added to the partial solution $ S$, then $ {s}_{v}=0 $ will become $ {s}_{v}=1 $.
\item \textbf{Action Space:} An action selects a vertex on the graph which is not included in the partial solution $ S$ and connected to the partial solution. We use the $ {S}^{*} $ set to represent in section \ref{greedy}. The purpose of selecting adjacent vertices is to maintain the tree's shape, which can avoid loops and unconnected vertices in the partial solution.
\item \textbf{Rewards:} The reward depends on the current state and action. When adding a vertex $ v$ moves to the next state $ S':= (S,v) $, there are two vertices: terminal and non-terminal. Our goal is to find all terminals with the shortest path length. So when the added vertex is non-terminal, the reward function $ r(S,v) $ depends on the change of the cost function $ C(S,G) $ and the remaining path weight from terminals $|x_v|$ defined in Section \ref{Initial}. When the added vertex is the terminal, the reward function is positive to distinguish the non-terminal, 
{adding a positive number.} The Equation(\ref{eq:rs}) is as follows: 
\begin{equation}
r(S,v) = \left\{ \begin{array}{l}
C(S',G) - C(S,G) - |x_v|, v \notin T\\
C(S',G) - C(S,G) - |x_v| + c, v \in T
\end{array} \right.\
\label{eq:rs}
\end{equation}\\
where $c$ depends on the weight of edges. 
{Different types of graphs have weights of different magnitudes. For edges with large weights, if the given $c$ is small, it will not be effective.}
\item \textbf{Policy:} From section \ref{greedy}, we adopt a deterministic greedy policy in the following form Equation(\ref{eq:policy}).
\begin{equation}
\pi (v|S) = \arg {\max _{v \in {S^*}}}Q(S,v) 
\label{eq:policy}
\end{equation}
Such a greedy strategy can easily cause the final network model to fall into a locally optimal, so we set the exploration rate $ \varepsilon $ usually 0.1, i.e., 
{the probability of 0.9 chooses the greedy algorithm, the probability of 0.1 randomly chooses $ v \in {S}^{*} $.} 
\item \textbf{Termination:} All terminals are added to the partial solution $ S$, i.e. $ T \in S$. 
\end{itemize}

\begin{figure*}[!htb]
\centering
\includegraphics[width=\linewidth]{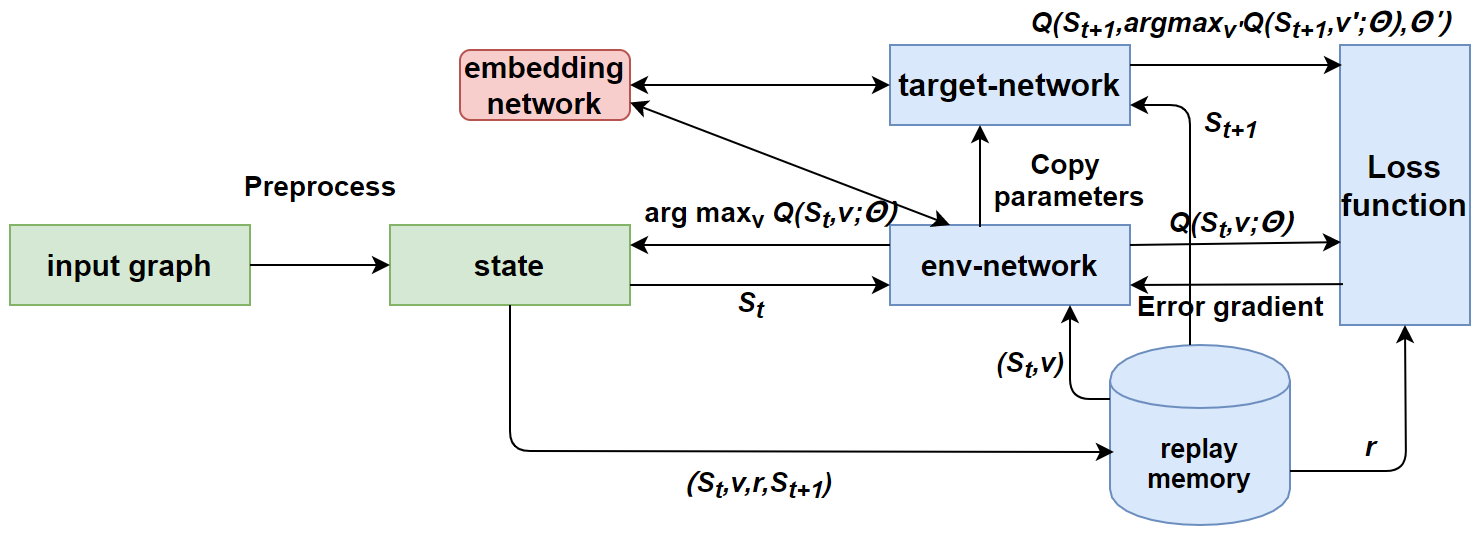}
\caption{Learning model of DDQN for evaluation function $Q$. The structure of the env-network and the target-network comes from the embedding network proposed in Section \ref{sec:embedding}.}
\label{fig:DQN}
\end{figure*}

We use standard Q-learning to update the network parameters, and each step uses the SGD optimizer to  minimize the square loss: 
\begin{equation}
J(\theta ) = {(y - Q({S_t},{v_t};\theta ))^2}   
\end{equation}
where $ y$ is target-network with the same structure but different parameters. According to the definition of deep Q networks (DQN) \cite{mnih2015human,silver2017mastering}, the target-network will have the following formula for estimation:

\begin{equation}
y =\left\{ \begin{array}{l}
r({S_t},{v_t}), terminal \\
\gamma {\max _{v'}}Q({S_{t + 1}},v';\theta ') + r({S_t},{v_t}), otherwise

\end{array} \right.\
\end{equation}

where $ \gamma $ represents the discount rate (0 $ \sim $ 1), which indicates the attenuation from the next state information. $\theta'$ is the parameter of the target network. The $max$ operation used in the above DQN makes the evaluation of an action value overestimate. Especially for NP-hard problems, there are strict requirements on accuracy, and a slight deviation may lead to unsatisfactory final results. Based on DQN, the overestimation is eliminated by decoupling the vertex selection and the calculation of the target-network. In DDQN, we do not directly find the maximum $Q$ value of each vertex in the target-network, but first, find the vertex corresponding to the maximum $Q$ value in the env-network. As shown in the following formula:
\begin{equation}
v'=\underset{v'}{\arg\max}Q(S_{t+1},v';\theta)
\end{equation}
then we substitute this $v'$ into Equation (10) to obtain a new target-network estimate:
\begin{equation}
y=\gamma Q(S_{t+1},\underset{v'}{\arg\max}Q(S_{t+1},v';\theta);\theta')+r(S_{t,}v_t)
\end{equation}

The learning process of the evaluation function $ Q$ is illustrated in Fig. \ref{fig:DQN}. In the training process, two additional vital ideas are introduced that are experience replay memory and target-network. The correlation of continuous samples will make the variance of the parameter update more extensive, and experience replay memory reduces this correlation by randomly sampling in the memory buffer. The parameters of the target-network remain unchanged for a period time, which reduces the correlation between the env-network and the target-network and improves the stability of the algorithm.



\section{Evaluation}
\label{sec:evaluation}

\subsection{Experimental environments}
All the experiments were executed on a single computer with the following specification: CPU: Inter(R) Core(TM) i5-8400 2.80GHz, GPU: NVIDIA GeForce GTX 1660 Ti 6GB, RAM: 16GB, Deep learning framework: python3.7 and pytorch 1.4.0, CUDA: 101, Operation System: Windows 10.
\subsection{Datasets introduction}
Due to the wide variety and complex structure of STP, we use both synthetic instances and real-world instances to test the performance of \system{}. The synthetic instances are affected by the generation rules and will be kept in the same distribution as possible. Real-world instances are often used as benchmarks in traditional methods. 
\begin{itemize}
\item \textbf{Synthetic instances}: To evaluate the effectiveness of \system{}, we generate Random-Regular (RR) \cite{molloy1995a}, Erdos-Renyi (ER) \cite{erdHos1960evolution} and Watts-Strogatz (WS) \cite{watts1998collective} graphs which are often used in various graph combination optimization problems, such as MVC, MAXCAT. The unique structure of STP is that the number and location of terminals will affect the distribution of generated graphs. To select terminals, we define a ratio  $m$ between the number of vertices and the number of terminals, e.g., every vertex has the probability of $m$ being a terminal. We attach the weight of each edge in the range of $[1,n]$, where n is a constant. Random graph generation and vertex selection ensure the complexity of generated instances, making it sufficient to verify the stability of \system{}.
\item \textbf{Real-world instances}: We use the standard SteinLib Testsets \cite{koch2001steinlib} that are publicly available. The instances in SteinLib have been used as benchmarks for STP. We target different types of graphs to check the applicability of \system{}. The following four types of graphs will be used in our experiments: sparse with random weights, VLSI applications, Rectilinear graphs and Group Steiner Tree Problems.
\end{itemize}

\subsection{Comparison methods}

\begin{itemize}
\item \textbf{Classic}: STP can be approximated by computing the minimum spanning tree of the subgraph induced by the terminal nodes. Given a minimal spanning tree, we can construct a subgraph by replacing each edge in the tree with its corresponding shortest path in a complete graph. We compare it as a classic algorithm\cite{KouA} that usually appears as a baseline in heuristic algorithms.
\item \textbf{\system{}}: \system{} uses the combination of graph embedding and DDQN, learns to select the optimal vertex in the current state. For stabilizing the training, we set the exploration probability from 0.1 to 0 in a linear way. The selection of hyperparameters is trained through small generation instances and fixed for other instances.
\item \textbf{GNNs}: In the previous introduction, we have explained why the current GNNs are not suitable for our problem. To verify its correctness, we select several representative GNNs as a comparison of {\system{}}. The original GNNs are mainly used in node classification or other graph combination optimization problems. To handle STP, we use the same encoder network and decoder network as {\system{}}. To reflect the difference, we use GNNs instead of the processor network that we proposed in {\system{}}. Finally, we use the same reinforcement learning for training. 
\begin{itemize}
\item\textbf{S2V}: S2V\cite{khalil2017learning} is used in MVC, MC, TSP, and other issues to capture the structural information of the graph and has achieved amazing results.
\item\textbf{GCN}: GCN is the most common baseline for node classification and node prediction. Since the efficient performance on the graph, it is also used in many NP-hard problems, such as MIS, MVC, etc\cite{mittal2019learning,li2018combinatorial}.  
\item\textbf{GAT}: Graph Attention Networks (GAT)\cite{velikovi2020neural,velickovic2018graph} adds an attention mechanism so that the network will first capture more similar adjacent nodes, thereby improving the model's accuracy. \item\textbf{MLP}: This part does not use GNNs, and only adds a few Multi-Layer Perceptrons (MLP) to improve the generalization ability of the encoder network. 
\end{itemize}
\end{itemize}

\subsection{General Parameter Tuning of \system{}}
\textbf{Experiment Settings.}
{In order to evaluate the performance of \system{} in different hyperparameters, model training was performed on RR graphs with 30 vertices.} We train the following four more important hyperparameters and observe their convergence curves. First, the test of the model with \{8, 16, 32, 64, 128\} batch size. Second, in the learning rate test, we measure the performance of \{$10^{-3}$,$10^{-4}$,$10^{-5}$\}. Third, as for the choice of K, we have explained the meaning of K in Section \ref{Initial}, so our experiments test \{1, 2, 3, 4\} to evaluate the influence. Finally, the discount rate $ \gamma $ is an important hyperparameter in reinforcement learning and generally set to 0.8 $\sim$ 0.9. But 
{in our experiment, the result shows that} big $ \gamma $ cannot give the model a better performance. Therefore we test three levels of \{0.2, 0.4, 0.8\}.\\
\begin{figure}[!htb]
\centering
\subfigure[Batch Size]{
\begin{minipage}{0.45\columnwidth}
\centering
\includegraphics[width=\linewidth]{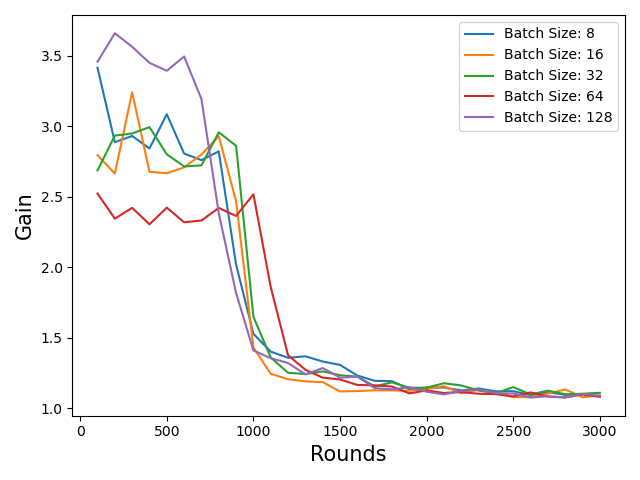}
\end{minipage}
}
\subfigure[K]{
\begin{minipage}{0.45\columnwidth}
\centering
\includegraphics[width=\linewidth]{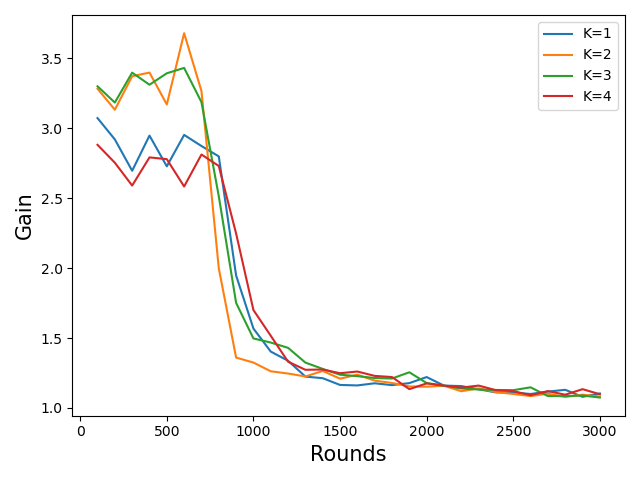}
\end{minipage} 
}
\subfigure[Learning Rate]{
\begin{minipage}{0.45\columnwidth}
\centering
\includegraphics[width=\linewidth]{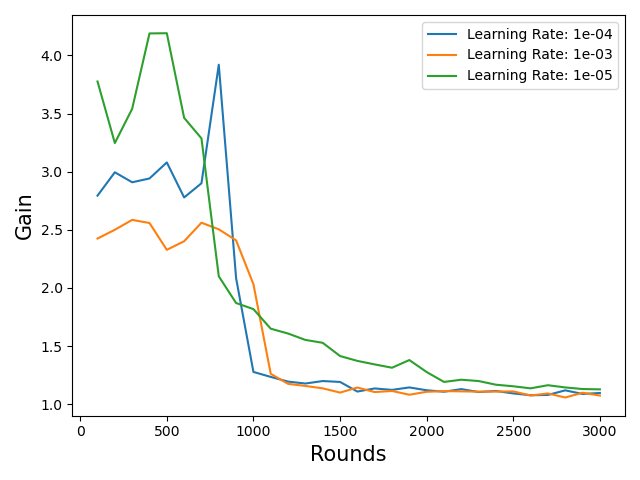}
\end{minipage} 
}
\subfigure[$\gamma$]{
\begin{minipage}{0.45\columnwidth}
\centering
\includegraphics[width=\linewidth]{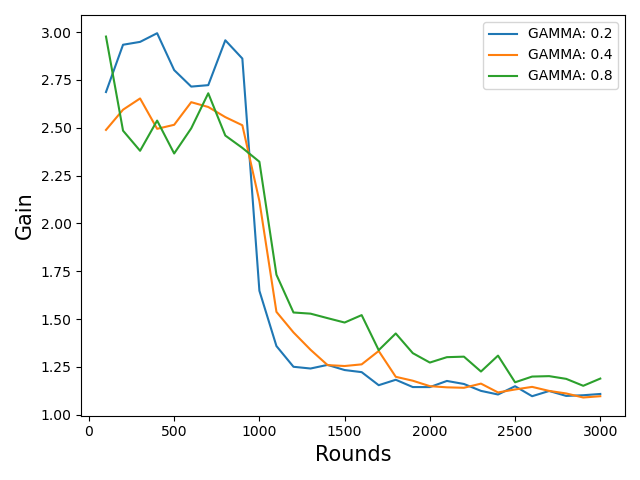}
\end{minipage} 
}
\caption{Comparison of different hyperparameters in \system{}.}
\label{tuning}
\end{figure}\\
\textbf{Metrics. }In our experiments, a round means learning steps from the beginning to the end of STP on a single graph \cite{khalil2017learning}, and $Gain$ represents an approximate ratio as set in Section \ref{sec:generation}. 
{For each training, we change one of the hyperparameters and fix others to observe the impact on the performance.}\\
\textbf{Findings. } 
In Fig. \ref{tuning}, we find no apparent difference between different batch sizes, even if their initial gains are different. So overall considering the cost of calculation and speed of convergence, batch 16 is the most suitable condition. Similarly, setting the $K$ number to 2 can give the whole model the best performance both in calculating quantity and dropping rate. As shown in Fig. \ref{tuning}(c), when the learning rate is small or big, $Gain$ drops slowly. A learning rate of 1e-04 results in the best trade-off. The most apparent  result to emerge from Fig. \ref{tuning}(d) is that $\gamma=0.2$ is the optimal selection. 

\subsection{Comparison of solution quality in generated instances}
\label{sec:generation}
\textbf{Experiment Settings.}  We mainly compare \system{} to several GNNs modified to suit STP on three different types of generated graphs. Since the generated graphs lack the optimal solution labels, we use the Classic result as the reference standard. We set 30, 50, and 100 vertices on three types of graphs, $m=0.2$. On the generated graphs, we augment each edge with a weight drawn uniformly at a random integer from $[1,5]$. We train 6000 rounds on graphs of 30 and 50 vertices for each category, but 4000 rounds on graphs of 100 vertices. It is learned from experimental experience that for each graph with 100 vertices, the model needs to take many steps to complete a round, and more information will be learned, so the convergence tends to stabilize after 4000 steps. The comparison methods keep the same setting as \system{}.\\
\textbf{Metrics.} We use an approximate ratio between \system{} and the Classic approach to evaluate the solution quality on generation instances. It is defined as $Gain(S,G) = \frac{{C(S,G)}}{{Classic(G)}}$, where $C(S,G)$ is the solution value of \system{} or other GNNs. $Classic(G)$ is the solution value of the Classic approach on graph $G$. A lower approximation ratio is better. To reduce the error caused by the randomness of the generated graphs, we report the average results of 200 generated graphs, and then take the optimal value of the five average results as the final result for comparison. \\

\begin{table*}[htbp]
\centering
\caption{Results on the generation instances}
\resizebox{\textwidth}{!}{
\begin{tabular}{cccclccclccccccl}
\hline
\multirow{2}{*}{Name} & \multicolumn{3}{c}{RR} &  & \multicolumn{3}{c}{ER} &  & \multicolumn{3}{c}{WS} &  & \multicolumn{3}{c}{Fixed Weight 50} \\ \cline{2-4} \cline{6-8} \cline{10-12} \cline{14-16} 
 & 30 & 50 & 100 &  & 30 & 50 & 100 &  & 30 & 50 & 100 &  & RR & ER & \multicolumn{1}{c}{WS} \\ \cline{1-4} \cline{6-8} \cline{10-12} \cline{14-16} 
MLP & \textbf{1.021} & \textbf{1.022} & \textbf{1.033} &  & 1.048 & 1.111 & 1.349 &  & 1.050 & 1.047 & 1.044 &  & 0.959 & 0.976 & 0.959 \\
GAT & 1.314 & 1.303 & 1.137 &  & 1.055 & 1.148 & 1.364 &  & 1.057 & 1.044 & 1.045 &  & 0.953 & 1.124 & 0.957 \\
GCN & 1.313 & 1.358 & 1.277 &  & 1.120 & 1.212 & 1.463 &  & 1.087 & 1.074 & 1.076 &  & 1.011 & 1.148 & 0.985 \\
S2V & 1.083 & 1.153 & 1.062 &  & 1.075 & 1.287 & 1.445 &  & 1.113 & 1.058 & 1.050 &  & 0.950 & 0.974 & 0.958 \\
\system{} & 1.024 & 1.024 & 1.035 &  & \textbf{1.027} & \textbf{1.085} & \textbf{1.308} &  & \textbf{1.023} & \textbf{1.019} & \textbf{1.023} &  & \textbf{0.942} & \textbf{0.966} & \textbf{0.951} \\ \hline
\end{tabular}}
\label{table1}
\end{table*}

\textbf{Findings.} We begin by reporting results on the RR graphs. Table \ref{table1} shows that all three ratios of MLP are the lowest on the RR graphs. \system{} is almost the same as the optimal result. However, the results of GAT and GCN are terrible, probably because they failed to capture useful local information, and ignored the initial vertex weights. As for ER graphs and WS graphs, we can notice that the results of \system{} are significantly better than other methods. In most cases, \system{} controls the approximate ratio below 1.04. Besides, we see that the performance of GNNs is not bad, especially GAT, which is better than MLP on some graphs. The Comprehensive analysis found that the reasons for the different results may be related to the rules of graph generation. The RR graphs will maintain the same degree when they are generated, so the graph structure will be simple so that GNNs cannot capture effective local information. Although the generation rules of ER graphs and WS graphs are complex enough to model many real-world networks, the performance of GNNs compared to MLP is not significantly improved. The results also confirm our previous hypothesis that the local information captured by GNNs is not suitable for STP. Lastly, we perform an additional study. To reduce the influence of the edge weight, we set a fixed weight of 1. Noting that \system{} has a ratio of less than 1 on the three types of graphs, and the performance of MLP and S2V is also satisfactory, indicating that our encoder network and decoder network play an important role.

\subsection{Generalization to larger instances}
\textbf{Experiment Settings.} This part is to investigate the generality of \system{}. Graph embedding uses the same parameters that can ensure the model to train and test on graphs of different sizes. We train \system{} on graphs with 30 vertices, and then retain its parameters and generalize it to larger graphs with up to 150 vertices. For each category of generated graphs, we choose the most competitive method by observing the results in Table \ref{table1} for comparison. \\
\textbf{Metrics.} We continue to adopt the same evaluation indicators and training as section \ref{sec:generation}.
\begin{table}[!htbp]
\centering
\caption{The results on generalization}
\begin{tabular}{cccccc}
\hline
\multicolumn{2}{c}{Test Size} & 50 & 80 & 100 & 150 \\ \hline
\multirow{2}{*}{RR} & \system{} & 1.029 & 1.032 & 1.044 & 1.068 \\ \cline{2-2}
 & MLP & 1.040 & 1.092 & 1.131 & 1.332 \\ \hline
\multirow{2}{*}{ER} & \system{} & 1.133 & 1.279 & 1.372 & 1.522 \\ \cline{2-2}
 & GAT & 1.169 & 1.341 & 1.427 & 1.566 \\ \hline
\multirow{2}{*}{WS} & \system{} & 1.035 & 1.048 & 1.048 & 1.069 \\ \cline{2-2}
 & GAT & 1.061 & 1.083 & 1.099 & 1.201 \\ \hline
\end{tabular}
\label{table2}
\end{table}\\
\textbf{Findings.} As we demonstrate in Table \ref{table2}, the results further solidify the outperformance of \system{}. For each category and size of graphs, the $Gain$ of \system{} is lower than its most competitive comparison method. It is worth noting that MLP, which performed well in previous experiments, is far inferior to our approach in generalization. The reason may be that the MLP can easily cause overfitting. On the contrary, we observe that \system{} has no significant overfitting from the results, and the $Gain$ can be controlled below 1.08 in the RR graphs and WS graphs with up to 150 vertices.

\subsection{Experiments on real-world instances}
\textbf{Experiment Settings.} In addition to the experiments for synthetic data, we verify the reliability of \system{} in four categories real-world instances with Classic and GNNs. Each real-world instance provides an public optimal solution, which helps measure experimental results. Instead of training and testing on many graphs, we use Active Search\cite{bello2016neural} that actively updates its parameters while searching for candidate solutions on a single instance. Remarkably, it can also produce a satisfactory solution from an untrained model. The best results during training will be stored.\\
\textbf{Metrics.} The approximation ratio of solution $S$ to a instance $G$ is defined as $R(S,G) = \frac{{C(S,G)}}{{OPT(G)}}$, where $C(S,G)$  is the optimal solution found by our model or comparison method, $OPT(G)$ is the optimal solution provided by SteinLib website.\\
\begin{table}[htbp]
\centering
\caption{The results on real-world instances} 
\resizebox{\columnwidth}{!}{
\begin{tabular}{cccccccc}
\hline
Name & OPT & \system{} & MLP & GCN & GAT & S2V & Classic \\ \hline
b02 & 83 & \textbf{86} & \textbf{86} & 89 & 87 & 91 & 90 \\
b03 & 138 & 144 & 147 & 144 & 150 & 148 & \textbf{140} \\
b04 & 59 & 63 & 62 & 67 & 66 & 71 & \textbf{59} \\
b05 & 61 & \textbf{62} & \textbf{62} & \textbf{62} & \textbf{62} & 64 & 64 \\
b10 & 86 & \textbf{90} & \textbf{90} & 94 & \textbf{90} & \textbf{90} & 98 \\
b11 & 88 & \textbf{92} & 96 & 95 & \textbf{92} & 96 & 93 \\
b13 & 127 & \textbf{133} & 134 & 143 & 139 & 139 & 137 \\
b16 & 165 & \textbf{165} & 169 & 169 & 169 & 169 & 175 \\
lin01 & 503 & \textbf{503} & \textbf{503} & \textbf{503} & \textbf{503} & 897 & \textbf{503} \\
lin02 & 557 & 559 & 559 & 561 & 559 & 1112 & \textbf{557} \\
lin03 & 926 & \textbf{926} & 926 & 929 & 929 & 1105 & 932 \\
lin04 & 1239 & \textbf{1239} & \textbf{1239} & 1383 & 1465 & \textbf{1239} & 1267 \\
lin05 & 1703 & \textbf{1733} & 1787 & 2097 & 1937 & 1813 & 1808 \\
lin06 & 1348 & \textbf{1382} & 1414 & 1612 & 1608 & 1514 & 1412 \\ 
\multicolumn{1}{l}{es10fst01} & 22920745 & \textbf{22920745} & \textbf{22920745} & \textbf{22920745} & \textbf{22920745} & \textbf{22920745} & 23090747 \\
es10fst03 & 26003678 & \textbf{26003678} & \textbf{26003678} & \textbf{26003678} & \textbf{26003678} & 26496603 & 28476616 \\
wrp3-11 & 1100361 & \textbf{1100516} & 1100556 & 1100561 & 1100548 & 1100578 & 1700323 \\
wrp3-12 & 1200237 & \textbf{1200305} & 1200318 & 1200319 & 1200325 & 1400319 & 1900155 \\ \hline
R & 1 & \textbf{1.019} & 1.027 & 1.064 & 1.056 & 1.094 & 1.103 \\ \hline
\end{tabular}}
\label{table3} 
\end{table}\\
\textbf{Findings.} Table \ref{table3} shows that \system{}  has a good approximate ratio for four sets of real-world instances. We noticed that compared with the Classic algorithm, \system{} is improved by 7.5$\%$. The possible reason is that a single graph is easier to understand by the network, without considering issues such as robustness. Therefore, the MLP method that performs poorly in generalization tests can also achieve good results. However, GNNs pay too much attention to local features, resulting in the final results easily falling into local solution. In summary, almost all deep learning methods control the ratio below 1.1, which verifies the rationality of our reinforcement learning settings.

\subsection{Reducing other NP-hard problems to STP }
\textbf{Experiment Settings.} We transformed three widely used NP-hard problems (SAT, MVC, and X3C) into STP. The SAT instances come from SATLIB \cite{hoos2000satlib}. We use a data set called 'uf20-91', which contains 1000 instances. Each instance is satisfiable and has 20 variables, 91 clauses. The MVC and X3C instances are synthetic. Each MVC instance contains 30 vertices and follows the generation rules of the WS graph. Each X3C instance contains 6 variables and 36 clauses. Every instance will be converted into a corresponding STP instance with a bound. If the cost of finding a path on STP is less than that of a bound, we can find a corresponding solution to the original problem. \\
\textbf{Metrics.} Similar to the previous metric, we set an approximation ratio between STP solution and bound, which is defined as $B(S,G) = \frac{{C(S,G)}}{{Bound(G)}}$, where $C(S,G)$  is the optimal solution found by our model or comparison method, $Bound(G)$ is the bound. We take the average ratio of 200 instances as the results.
\begin{table}[htbp]
\centering
\caption{The results on transformed instances}
\begin{tabular}{llllll}
\hline
Name & \system{} & MLP & GAT & GCN & S2V \\ \hline
SAT & 1.035 & \textbf{1.022} & 1.029 & 1.026 & 1.139 \\
MVC & \textbf{1.105} & 1.130 & 1.191 & 1.1386 & 1.190 \\
X3C & \textbf{1.083} & 1.088 & 1.102 & 1.090 & 1.098 \\ \hline
\end{tabular}
\label{table4}
\end{table}\\
\\
\textbf{Findings.} From the Table \ref{table4}, most models can obtain approximate ratios very close to $B(S,G)$ on SAT instances, but still difficult to reach the bound of transformation. in addition, for MVC instances, the results are not satisfactory. Through analysis, it is found that reducing MVC instances to STP instances will result in a large number of edges joining (\eg MVC with 30 vertices and 60 edges to STP with 90 vertices and more than 4000 edges). Deep learning models cannot handle this situation well. But \system{} still achieved the best results.
\\

\subsection{Trade-off between running time and approximate ratio}
\textbf{Experiment Settings.} To find the trade-off between the running time and the approximate ratio of methods, we tested 100 RR graphs with 150 vertices. At the same time, considering the generalization, we trained on RR graphs with 30 vertices. To ensure fairness, all methods use the same CPU for testing. \\
\textbf{Metrics.} For the approximate ratio of the ordinate, we use the same settings as section \ref{sec:generation}. The abscissa represents the test time of a single graph.
\begin{figure}[htbp]
\centering
\includegraphics[width=\columnwidth]{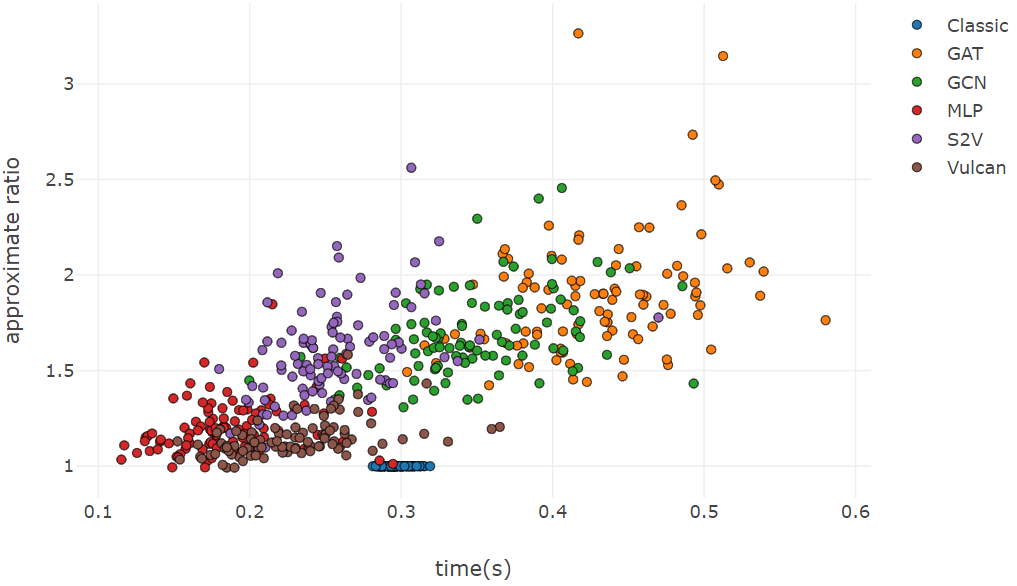}
E\caption{Time-approximation trade-off for RR graphs }
\label{time}
\end{figure}
\\
\textbf{Findings.} In Fig. \ref{time}, each dot
represents a solution found for a single problem instance. MLP is faster but unstable. In contrast, \system{} is slightly slower than MLP but maintains a lower ratio. We can also see that most instances of \system{} are faster than Classic. However, since the generalization of the model, the ratio is not up to Classic. Other GNN models are not competitive. Of course, \system{} can accelerate calculations on large-scale graphs by GPU, which is also one of the advantages of deep learning compared to heuristic algorithms. \\
\section{Conclusion}
\label{sec:conclusion}
We have presented an approach to solving STP with novel GNN and deep reinforcement learning. The core of \system{} is a novel, compact graph embedding that transforms high-dimensional graph structure data into a low-dimensional vector representation. \system{} greatly reduces the manual design, compared to the complex heuristic methods. \system{} can also find solutions to a family of NP-hard problems (e.g., SAT, MVC and X3C) by reducing them to STP. Experiments show that both in synthetic and real-world datasets, \system{} has achieved satisfying results. We see the presented work as a step towards a new family of solvers for NP-hard problems that leverage both deep learning and reinforcement learning. We will release the codes to support future progress along this direction.


%
%


%
%
\bibliography{vulcan}

\end{document}